%% file: main.tex
\documentclass[hidelinks]{isprs}
\usepackage{subfigure}
\usepackage{import}
\usepackage{setspace}
\usepackage{geometry} 
\usepackage{epstopdf}
\usepackage[labelsep=period]{caption}  

\usepackage{hyperref}
\usepackage{graphicx}
\usepackage[export]{adjustbox}
\usepackage{amsmath,amsfonts,amssymb}
\usepackage[acronyms]{glossaries}
\usepackage[sort]{natbib} 
	\renewcommand{\refname}{REFERENCES} 
\usepackage{color, soul}
\usepackage{diagbox}	
\usepackage{enumitem}
\usepackage{gensymb}
\usepackage[nameinlink, capitalize]{cleveref}
\usepackage{hhline}
\usepackage{inputenc}
\usepackage{todonotes}
\usepackage{algorithm}
\usepackage[noend]{algpseudocode}
\usepackage{float}
\usepackage{cuted} 
\usepackage[percent]{overpic}
\usepackage{tabularx} 

\input{mathops.tex}
\input{macros.tex}

\begin{document}

\title{Real-time dense 3D Reconstruction from monocular video data captured by low-cost UAVs}

\author{
M. Hermann\thanks{Corresponding author}\ \ \textsuperscript{1,2}, B. Ruf\ \textsuperscript{1,2}, M. Weinmann\ \textsuperscript{2}}

\address{
\textsuperscript{1}Fraunhofer IOSB, Karlsruhe, Germany -\\ \{max.hermann, boitumelo.ruf\}@iosb.fraunhofer.de\\
\textsuperscript{2}Institute of Photogrammetry and Remote Sensing, Karlsruhe Institute of Technology (KIT), Karlsruhe, Germany -\\ \{max.hermann, boitumelo.ruf, martin.weinmann\}@kit.edu \\

}


\commission{II, }{YY} 
\workinggroup{II/4} 
\icwg{}   


\input{acronyms.tex}

\input{chapters/00_abstract.tex}

\maketitle

\glsresetall 

\input{chapters/01_introduction.tex}

\input{chapters/02_related_work.tex}

\input{chapters/03_methodology.tex}

\input{chapters/04_experiments.tex}

\input{chapters/06_conclusion.tex}


{\footnotesize 
	\begin{spacing}{0.5}
    \setlength{\bibsep}{0.8pt}
		\bibliography{biblio} 
	\end{spacing}
}

\end{document}

%% file: macros.tex
\usepackage{xspace}

\newcommand{\ie}{i.e.\ }

\newcommand{\eg}{e.g.\ }

\newcommand{\m}{\,m\xspace}

\newcommand{\km}{\,km\xspace}

\newcommand{\ms}{\,ms\xspace}

\newcommand{\s}{\,s\xspace}

\newcommand{\fps}{\,fps\xspace}

\newcommand{\snSGM}{SGM$^{sn}$\xspace}


%% file: acronyms.tex
\newacronym{ASG}{ASG}{{Average Shading Gradient}}

\newacronym[shortplural={CNNs}, firstplural={convolutional neural networks (CNNs)}, longplural={convolutional neural networks}]{CNN}{CNN}{convolutional neural network}
\newacronym{COTS}{COTS}{commercial off-the-shelf}
\newacronym[shortplural={CRFs}, longplural={{Conditional Random Fields}}]{CRF}{CRF}{{Conditional Random Field}}
\newacronym{CT}{CT}{census transform}

\newacronym{DLT}{DLT}{{Direct Linear Transformation}}
\newacronym{DoG}{DoG}{{Difference of Gaussian}}

\newacronym{EPnP}{EPnP}{{Efficient Perspective-n-Point}}

\newacronym{GPGPU}{GPGPU}{general purpose computation on a {GPU}}
\newacronym{GPS}{GPS}{{Global Positioning System}}
\newacronym{GTA}{GTA V}{Grand Theft Auto V}

\newacronym{ICP}{ICP}{{Iterative-Closest-Point}}
\newacronym{IMU}{IMU}{{Inertial Measurement Unit}}
\newacronym{INS}{INS}{{Inertial Navigation System}}

\newacronym{LIDAR}{LiDAR}{{Light Detection and Ranging}}
\newacronym{L1-rel}{$\text{L1-rel}$}{{relative $\text{L1}$-Norm}}
\newacronym{L1-abs}{$\text{L1-abs}$}{{absolute $\text{L1}$-Norm}}

\newacronym[shortplural={MRFs}, longplural={{Markov Random Fields}}]{MRF}{MRF}{{Markov Random Field}}

\newacronym{MVS}{MVS}{Multi-View-Stereo}

\newacronym{NCC}{NCC}{normalized cross correlation}

\newacronym{PCL}{PCL}{{Point Cloud Library}}

\newacronym{RANSAC}{RANSAC}{{Random Sampling Consensus}}
\newacronym{RMSE}{RMSE}{{Root Mean Square Error}}
\newacronym{MAE}{MAE}{{Mean Absolute Error}}

\newacronym[shortplural={ROIs}, longplural={regions of interest}]{ROI}{RoI}{region of interest}

\newacronym{SAD}{SAD}{sum of absolute differences}
\newacronym{SFM}{SfM}{{Structure-from-Motion}}
\newacronym{SGBM}{SGBM}{{Semi-Global Block Matching}}
\newacronym{SGM}{SGM}{{Semi-Global Matching}}
\newacronym{SLAM}{SLAM}{simultaneous localization and mapping}
\newacronym{SMDE}{SMDE}{{Self-supervised Monocular Depth Estimation}}
\newacronym{SSIM}{SSIM}{Structural Similarity}
\newacronym[shortplural={STNs}, longplural={Spatial Transformer Networks}]{STN}{STN}{{Spatial Transformer Network}}
\newacronym[shortplural={surfels}, longplural={surface-elements}]{surfel}{surfel}{surface-element}

\newacronym[shortplural={UAVs}, longplural={unmanned aerial vehicles}]{UAV}{UAV}{unmanned aerial vehicle}

\newacronym{WTA}{WTA}{winner-takes-it-all}

\newacronym{SDF}{SDF}{signed distance function}

\newacronym{TSDF}{TSDF}{truncated signed distance function}
\newacronym{NBV}{NBV}{Next-Best View}
\newacronym{ToF}{ToF}{time-of-flight}

%% file: chapters/00_abstract.tex
\abstract{   
    Real-time 3D reconstruction enables fast dense mapping of the environment which benefits numerous applications, such as navigation or live evaluation of an emergency. %
    In contrast to most real-time capable approaches, our approach does not need an explicit depth sensor. %
    Instead, we only rely on a video stream from a camera and its intrinsic calibration. %
    By exploiting the self-motion of the \gls*{UAV} flying with oblique view around buildings, we estimate both camera trajectory and depth for selected images with enough novel content. %
    To create a 3D model of the scene, we rely on a three-stage processing chain. %
    First, we estimate the rough camera trajectory using a \gls*{SLAM} algorithm. %
    Once a suitable constellation is found, we estimate depth for local bundles of images using a \gls*{MVS} approach and then fuse this depth into a global \acrshort*{surfel}-based model. %
    For our evaluation, we use $55$ video sequences with diverse settings, consisting of both synthetic and real scenes. %
    We evaluate not only the generated reconstruction but also the intermediate products and achieve competitive results both qualitatively and quantitatively. %
    At the same time, our method can keep up with a $30$\fps video for a resolution of $768\times448$ pixels. %
}

%% file: chapters/01_introduction.tex
\section{INTRODUCTION}
\label{sec:intro}
\sloppy

The widespread availability of inexpensive but versatile \glspl*{UAV} combined with open-source photogrammetry software such as COLMAP \citep{schoenberger2016mvs,schoenberger2016sfm}, has greatly simplified the offline creation of high quality 3D models from aerial imagery. %
However, due to the focus on the best possible quality and a mostly multi-stage pipeline, these methods are not suitable for real-time processing -- despite the rapidly increasing performance of modern hardware. %
Yet, real-time dense 3D mapping of the environment alleviates a number of applications, ranging from autonomous navigation of \glspl*{UAV} in three-dimensional space to estimating the impact of emergency events. %
Even for non-urgent cases, a coarse 3D reconstruction in real-time can be beneficial in order to avoid potential holes and artifacts in a subsequent and time-consuming highly accurate 3D reconstruction, thus avoiding an expensive secondary data acquisition. %

In this work, we present an approach for incremental real-time 3D reconstruction and model generation. %
To this end, we rely exclusively on monocular RGB video data, since the use of alternatives such as stereo cameras or LiDAR systems is not practical for low-cost \gls*{COTS} \glspl{UAV}. %
Both often do not meet the requirements for range or resolution and reduce the maximum flight time due to higher weight. %
In case of stereo cameras, the baseline of the individual cameras is usually too small to ensure adequate coverage of the scene depth. %
However, a disadvantage of our method can be seen in a more complex processing. %

Our approach for real-time 3D reconstruction consists of three main parts: %
\begin{enumerate}
\item First, we use a \gls*{SLAM} algorithm to estimate the trajectory of the camera and sample image bundles suitable for a subsequent dense image matching and depth estimation.%
\item Given these image bundles, we perform a multi-view dense image matching and depth estimation based on a real-time plane-sweep sampling and a subsequent semi-global optimization scheme in the second step of our pipeline. %
\item Lastly, the depth maps and the corresponding color images get fused into a joint point cloud from which a 3D model is extracted. %
\end{enumerate}

This paper is structured as follows: In \Cref{sec:related_work}, we briefly review several approaches to image-based 3D reconstruction, both for offline and online processing. %
We then describe our method in \Cref{sec:methodology} and address the three main components of our work. %
We evaluate our approach in \Cref{sec:eval} using two datasets and discuss our findings. %
Finally, we summarize our work in \Cref{sec:conclusion} and provide a brief outlook on future work. %

%% file: chapters/02_related_work.tex
\section{RELATED WORK}%
\label{sec:related_work}
\sloppy

From a high-level view, the process of image-based 3D reconstruction and model generation can be subdivided into three consecutive steps, namely: camera pose estimation, depth estimation and depth map fusion. %
As part of software frameworks for offline 3D reconstruction without real-time constraints, like 
\begin{strip}
    \includegraphics[width=\textwidth]{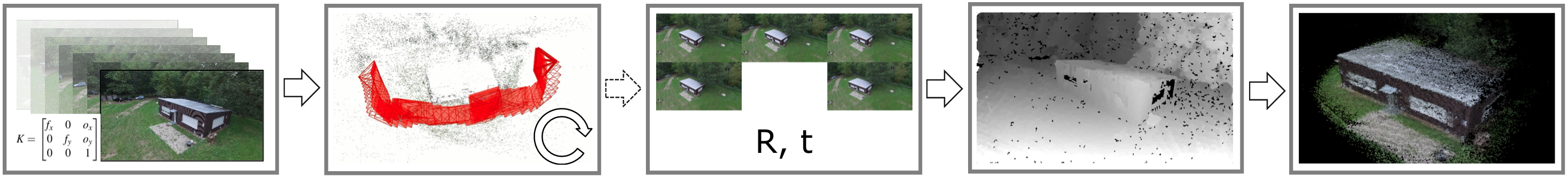}
    
    \small ~Input image sequence ~~~~~~~~~~~~~~~~~~~~ SLAM ~~~~~~~~~~~~~~~~~~Images $I_i$ and poses $\mathbf{P}_i$~~~~~~~ Estimate $D$ from $I_\mathrm{ref}$ ~~~~~~~~~Fusion using $I_\mathrm{ref}, D$ 
    \smallskip
    \captionof{figure}{Overview of our entire processing chain. %
    As input, we only need an image sequence and its intrinsic camera calibration. %
    Using a SLAM algorithm, we constantly track the current trajectory of the camera and select groups of five images $I_i$ and their poses $\mathbf{P}_i$,  suitable for depth estimation. %
    The estimated depth map $D$ is then fused into a global \gls*{surfel}-based model.}
    \label{fig:pipline_overview}
\end{strip}VisualSFM or COLMAP, this pipeline is typically instantiated by first computing a sparse reconstruction with the help of \gls*{SFM} \citep{schoenberger2016sfm, Wu2013towards, Frahm2010building} followed by a dense modeling with \gls*{MVS} \citep{schoenberger2016mvs, Furukawa2009accurate}. %

As part of a processing pipeline for real-time or online 3D reconstruction, such as presented in \citep{Kuhn2017tv, Pollefeys2008}, in which the 3D model is incrementally generated during the acquisition of the image data, the \gls*{SFM} pipeline is typically exchanged by a \gls*{SLAM} or an incremental \gls*{SFM} approach, which estimates the camera trajectory on-the-fly. %
While the full camera trajectory produced by a pure visual \gls*{SLAM} approach, like in \citep{orb_open_source, Engel2018}, is not as accurate as the trajectory computed by a global \gls*{SFM} pipeline, its precision within a confined window is sufficient for the subsequent \gls*{MVS} reconstruction. %
The process of dense modeling with \gls*{MVS} is usually subdivided into two separate steps, first dense image matching and depth estimation, second depth map fusion. %
When relying on an image sequence as input, the so-called plane-sweep algorithm, first introduced by \citet{Collins1996space}, allows for multi-image matching and depth estimation and was adopted by numerous studies for real-time dense depth estimation, such as \citep{Gallup2007, Sinha2014}. %
This requires a subsequent fusion of depth maps into a joint model as proposed in \citep{Kolev2014, Merrell2007}. %

In contrast, so-called Dense-\gls*{SLAM} approaches like presented in \citep{6696650} directly generate a dense model of the environment in real-time by %
using sensors that provide RGBD data and are usually based on stereo cameras, structured light or \gls*{ToF} sensors for measurement. %
Two particularly common methods for managing the resulting 3D model are as a voxel grid or as an unstructured point-based representation. %

With KinectFusion, \citet{izadi2011kinectfusion} have developed a voxel-based framework to reconstruct indoor scenes in real time using low-cost RGBD cameras such as the Microsoft Kinect. %
In their work, they extend the concept of the \gls*{SDF} from \citet{curless1996volumetric} by truncating the distance to the nearest surface, resulting in the so-called \gls*{TSDF}. %
In terms of regular voxel grids, the amount of memory required increases cubically with the resolution and size of the resulting model. %
To address this problem, there are various approaches, such as automatic \mbox{offloading} of inactive model regions \citep{Whelankintinuos}, the storage of voxels with varying resolution in hierarchical data structures \citep{zeng2012memory,fuhrmann2011fusion}, or the so-called voxel hashing \citep{niessner2013real} to store only populated voxels. %
However, they all have in common that they need additional algorithms such as ray-casting or marching cubes \citep{lorensen1987marching} in order to render individual sections from the voxel grid. %

As an alternative to volumetric approaches, \citet{pfister2000surfels} introduce the concept of \glspl*{surfel}, which extends the individual points in space with additional information such as normal vector, radius or confidence. %
Surfels have also been widely used for the task of Dense-\gls*{SLAM} \citep{keller2013real,Whelan2015ElasticFusionDS,Kolev2014}. %
A major drawback of this representation is its unstructured nature, which makes it difficult to examine the neighborhood of individual points and can lead to hole-ridden models. %
This can be addressed with spatial index structures and so-called splatting. %
Here, the individual points are rendered as ellipses with a given radius, normal vector and color \citep{rusinkiewicz2002real}. %

We extend our previous work on efficient depth estimation \citep{Ruf2019efficient}, which is based on a real-time multi-image matching with plane-sweep sampling, by combining it with the state-of-the-art ORB-\gls*{SLAM}2 algorithm \citep{orb_open_source} for the estimation of the camera trajectory and frame selection. %
Furthermore, we adopt the algorithm of \citet{Whelan2015ElasticFusionDS} for real-time fusion of depth maps in combination with color images. %
In this, we have chosen to rely on a surfel-based representation, due to its higher flexibility with respect to incremental model generation. %

%% file: chapters/03_methodology.tex
\section{METHODOLOGY}%
\label{sec:methodology}
\sloppy

As illustrated in \Cref{fig:pipline_overview}, the processing pipeline of our approach consists of three main parts: 1) Sparse mapping and estimation of the camera trajectory via \gls*{SLAM}. 2) Multi-view dense image matching and depth estimation. 3) Fusion of depth maps into a joint point cloud. %
In the following sections, we will go into more detail about the three main components of our processing pipeline. %

\subsection{Sparse mapping and estimation of the camera trajectory}
For rough camera tracking and estimation of the relative transformation between images, that are to be used for depth estimation, we rely on ORB-SLAM2, as it provides a good tradeoff between speed and accuracy \citep{mur2015orb}. %
ORB-SLAM2 consists of three main components:

\begin{itemize}
    \item Camera tracking and feature extraction
    \item Local mapping and local bundle adjustment
    \item Loop detection with subsequent global bundle adjustment
\end{itemize}

By parallelizing these individual tasks, ORB-SLAM2 achieves a global long-term tracking that is still lightweight enough to run on a standard CPU \citep{orb_open_source}. %
During the first stage, features are extracted from the current image to perform camera tracking. %
For this purpose, the so-called ORB features are used, which are particularly robust in terms of rotation and scaling \citep{orb_feature}. %
The extracted features are then used to estimate the position in comparison to the other images. %
The second component takes care of environment mapping and optimizations using local bundle adjustments. %
In addition, a procedure is used that detects loops and compensates for the drift detected by performing a global bundle adjustment. %
In case the tracking is lost, a relocalization approach based on binary bags of words is used \citep{DBoW2}. %
For the representation of the camera trajectory, ORB-SLAM2 uses a pose graph in which only certain frames are stored. %
These so-called keyframes act as support points and are optimized in downstream bundle adjustments. %
This leads to better performance, however, no updated information is available for frames that are not selected as keyframes. %

In our processing chain, we mainly use ORB-SLAM2 for estimating camera motion in coherent groups of spatially close images. %
By limiting ourselves to keyframes, we also take advantage of ORB-SLAM2's internal logic for selecting images that display novel content. %
In this way, we avoid processing redundant information in costly downstream steps such as depth estimation. %
Because we mainly use camera poses for our online processing that have not been globally optimized, we do not benefit from loop closures and subsequent global bundle adjustments. %
Due to changes in the sparse map of ORB-SLAM2, arising from these optimizations, the absolute poses across all frames are not consistent over time. %
However, to have a certain consistency between the poses of the images that are sampled for the subsequent depth estimation, we make sure they are part of a local bundle adjustment. %

By setting minimum and maximum thresholds for camera movement between frames, a search is performed for suitable frames from which depth has not yet been estimated. %
The threshold values are strongly dependent on the depth estimation method used. %
We use a maximum rotation of $5$ degrees as an empirically determined upper limit for the method used here. %
For translation, we set a minimal value of $0.02$ that we use across all experiments, but which is not interpretable due to the monocular nature of the videos. %
Once a group of five images $I_i$ that meet the respective criteria is found, they are passed to the depth estimation along with their camera poses $\mathbf{P}_i$. %
This process of constant tracking by ORB-SLAM2 and the search for suitable images for depth estimation is illustrated in the first three images in \Cref{fig:pipline_overview}. %


\subsection{Multi-view dense image matching and depth estimation}%

In the second stage of our processing pipeline, we perform a multi-view dense image matching and depth estimation. %
In this, we use the efficient hierarchical plane-sweep multi-image matching with a subsequent surface-aware semi-global cost volume optimization (\snSGM) proposed by \citet{Ruf2019efficient}. %
This approach takes five images $I_i$ of an input sequence, as well as the corresponding camera poses $\mathbf{P}_i$, which we get from the camera tracking in the previous pipeline stage, and computes a depth map $D$, as well as a normal map $N$ and a confidence map $C$ for the middle one of the five input frames, \ie the reference frame $I_{\mathrm{ref}}$. %
Typical for any \gls*{MVS} approach, it is important that the selected input images depict the same part of the scene and provide enough parallax in order to estimate the scene depth. %
Given the bundle of input images, the approach first computes a Gaussian image pyramid, which allows for a hierarchical processing and, in turn, a computationally more efficient image matching and depth estimation. %
Starting at the highest pyramid level, \ie the one with the smallest image size, the approach performs a two-stage depth estimation at each level in the pyramid, generating the depth, normal and confidence maps in a coarse-to-fine manner. %
The results produced at each pyramid level serve as a prior to the estimation at the next level. %
Since to this end we only use the depth maps for the real-time 3D reconstruction, in the following we will only discuss the process of depth estimation as part of the approach presented by \citet{Ruf2019efficient}. %

In the first stage of the depth estimation, a real-time plane-sweep sampling is employed in order to construct a three-dimensional cost-volume, holding the pixel-wise matching costs for all positions of the sampling plane. %
In this, a fronto-parallel plane, parameterized by its normal vector and its distance from the pose of the reference camera $\mathbf{P}_{\mathrm{ref}}$, is swept along the direction of its normal vector through the scene space from $d_{\mathrm{max}}$ to $d_{\mathrm{min}}$. %
At each position of the plane, the four matching images, two on each side of $I_\mathrm{ref}$, are projected into the view of the reference camera via the sampling plane and matched with $I_\mathrm{ref}$. %
\citet{Ruf2019efficient} evaluate the use of two similarity measures for the computation of the matching costs, namely a truncated and inverted version of the \gls*{NCC} and the Hamming distance of the \gls*{CT}. %
Since the latter one is computationally less expensive, we chose the \gls*{CT} as a similarity measure in the multi-image matching. %
The scene space is sampled in such a way, that two consecutive planes introduce a maximum pixel displacement of 1 pixel along the epipolar line, leading to an increase in the sampling density the closer the plane is to the reference camera. %
Depending on the size of the scene and vantage points of the cameras, this can lead to a high number of sampling positions and, in turn, a cost volume with a very high memory consumption. %
In order to remedy this, \citet{Ruf2019efficient} employ for each pyramid level except the highest one, a dynamic cost volume and use the pixel-wise estimate of the previous level as a depth prior for a local sampling around the corresponding estimate. %

Since the distribution of the pixel-wise minimum matching cost in such a raw cost volume is very noisy, a subsequent Semi-Global optimization scheme, as proposed by \citet{Hirschmueller2008}, is used in the second stage of the depth estimation in order to regularize the cost volume before extracting the depth map by selecting the pixel-wise \gls*{WTA} solution. %
As stated by \citet{Ruf2019efficient}, the initial \gls*{SGM} algorithm \citep{Hirschmueller2008} favors fronto-parallel surfaces. %
However, when considering oblique imagery, the scene mostly consists of slanted surfaces, which is why \citet{Ruf2019efficient} propose to use surface normals to shift the zero-cost transition within the path aggregation of the \gls*{SGM} optimization, in order to better account for slanted surfaces. %
We use this so-called \snSGM variant as part of the dense-image matching and depth estimation of our processing pipeline. %
In this, the normal map computed at the previous pyramid level of the hierarchical processing is used for the adjustment of the zero-cost transition.  %
At the highest pyramid level, where no normal map is available as a prior, the standard \gls*{SGM} algorithm is used. %

Furthermore, at each pyramid level, a $5\times5$ median filter is used to filter outliers and reduce the noise of the resulting depth map. %
By applying a \gls*{DoG} filter in the final step of this pipeline stage, we remove image regions with little textural information, assuming that in such areas the process of dense-image matching is prone to errors due to the reduced
\begin{strip}
    \begin{center}
        \centering 
        \setlength\tabcolsep{1pt} 
        \begin{tabular}{ccccc}
            \includegraphics[width=33mm]{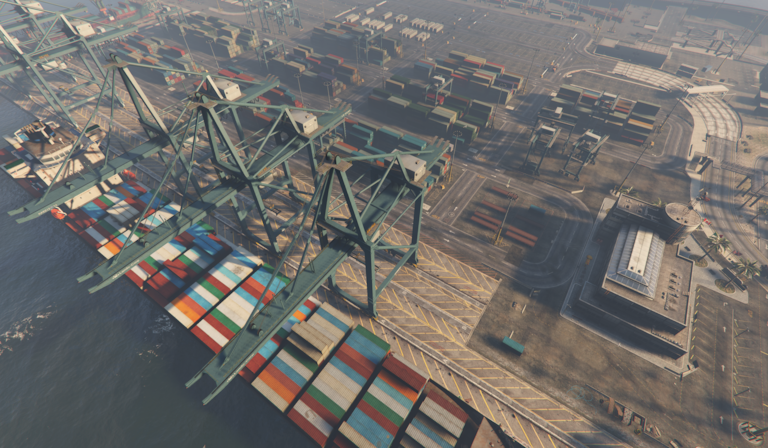}&
            \includegraphics[width=33mm]{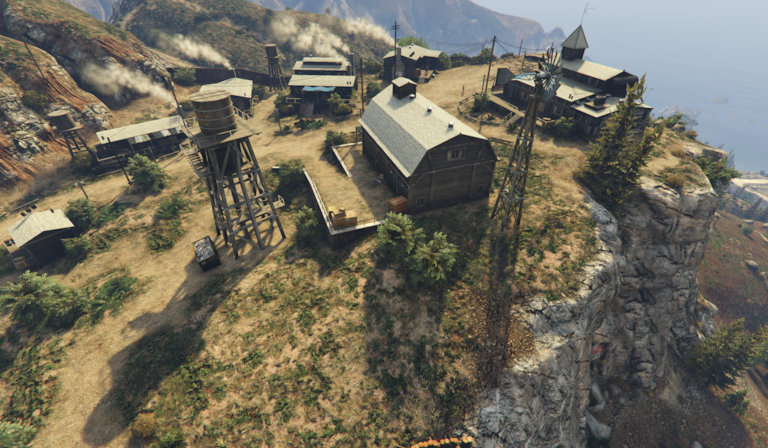}&
            \includegraphics[width=33mm]{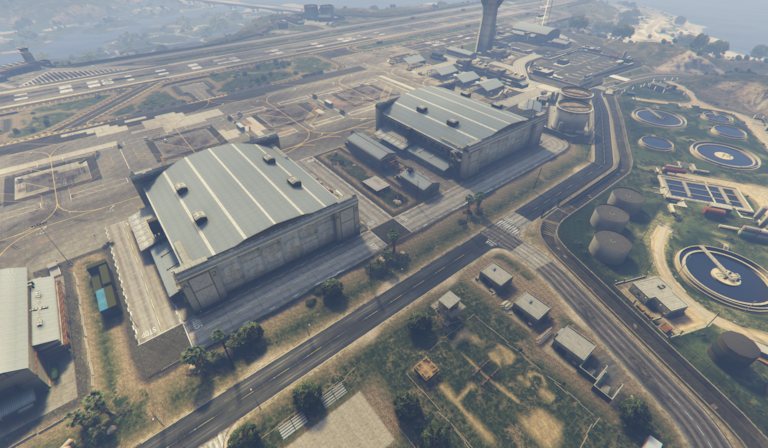}&
            \includegraphics[width=33mm]{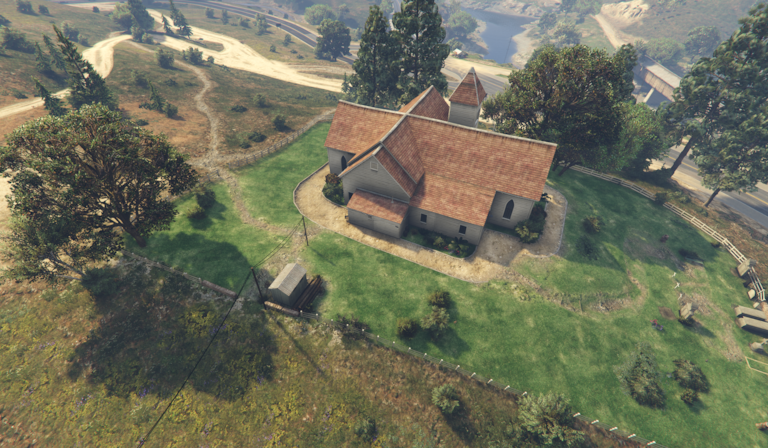}&
            \includegraphics[width=33mm,height=19.25mm]{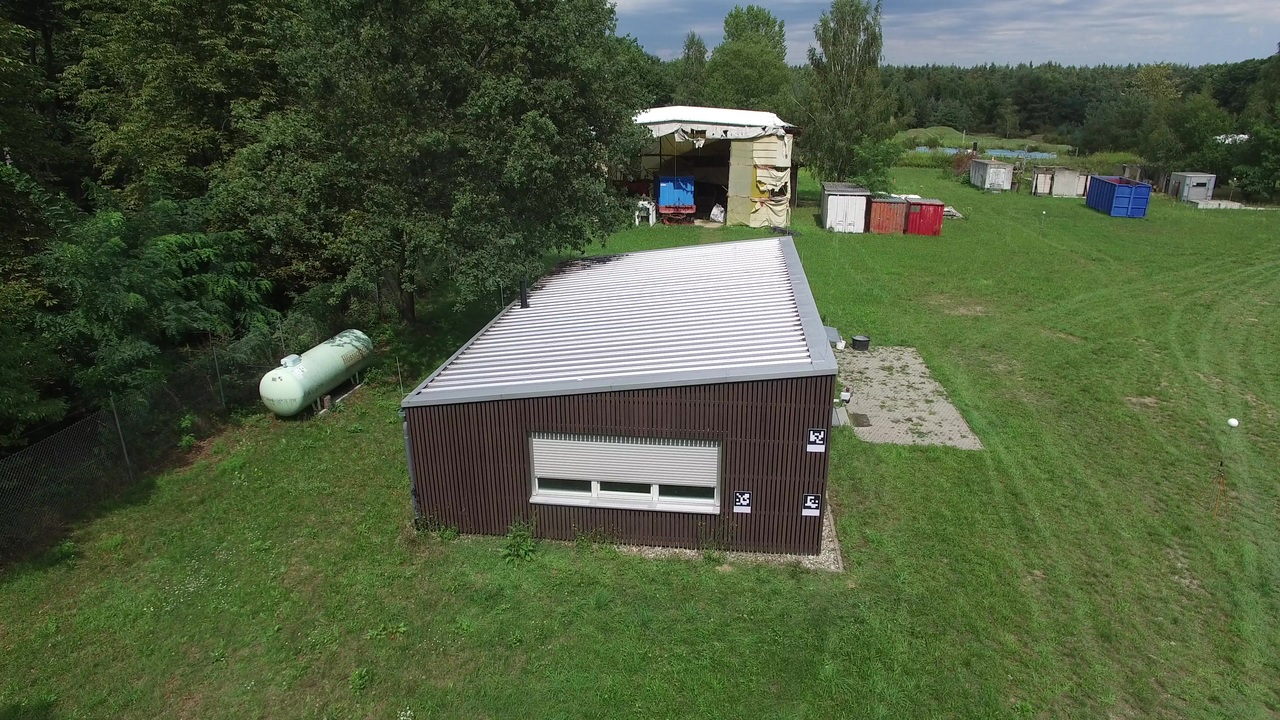}\\
            \includegraphics[width=33mm]{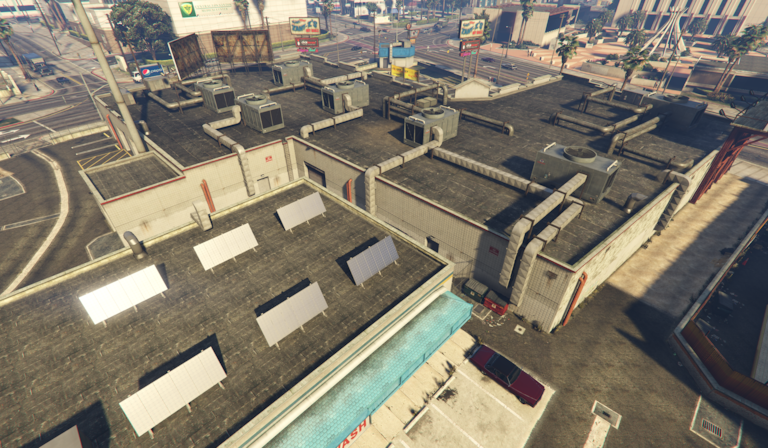}&
            \includegraphics[width=33mm]{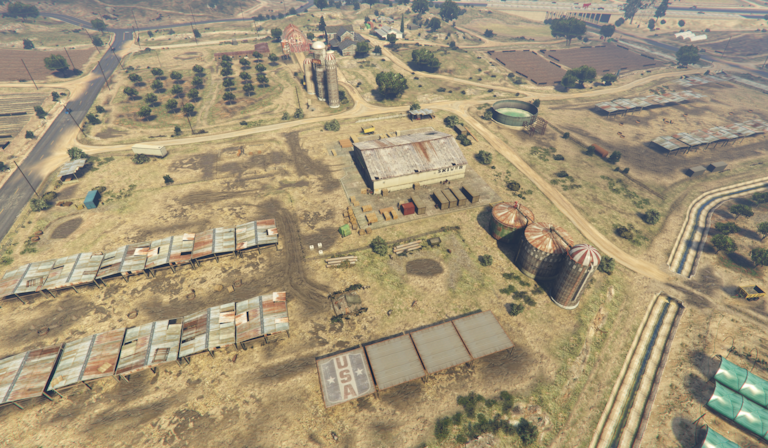}&
            \includegraphics[width=33mm]{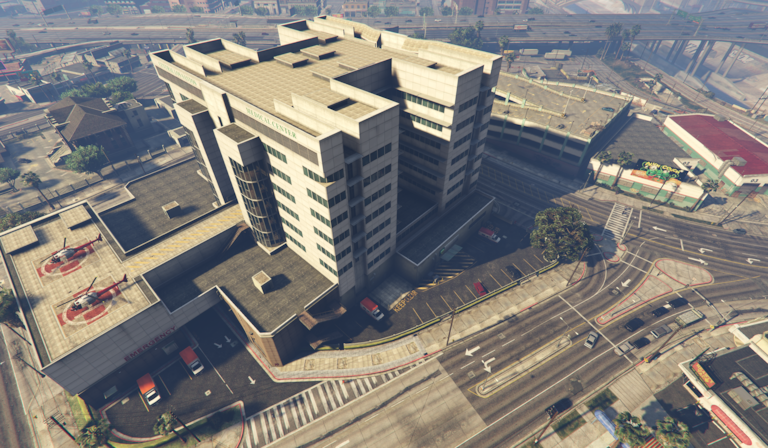}&
            \includegraphics[width=33mm]{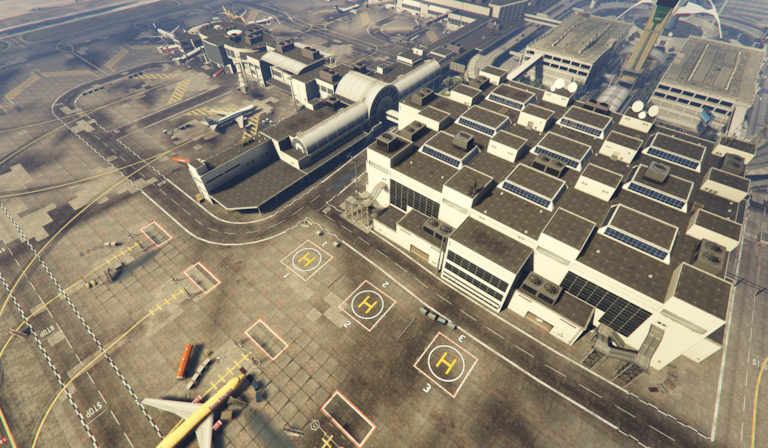}&
            \includegraphics[width=33mm,height=19.25mm]{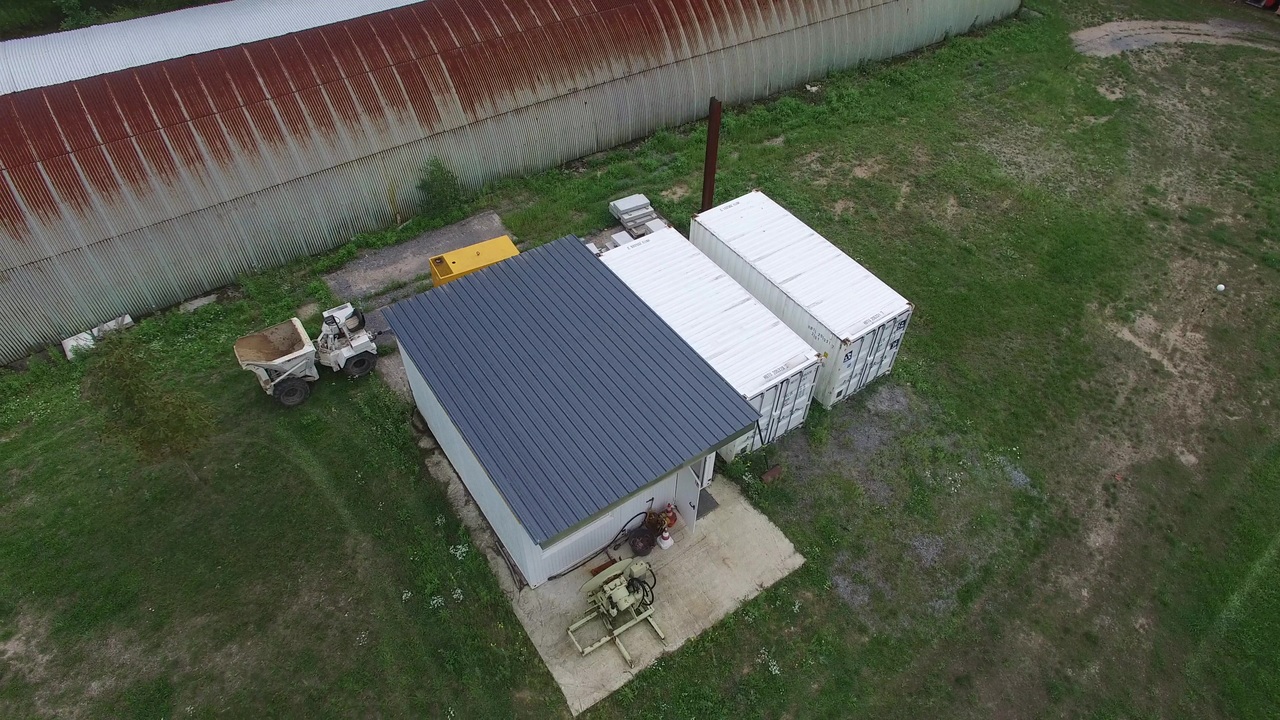}\\       
        \end{tabular}
        \small
        \captionof{figure}{Examples from our datasets. %
        The first four columns show scenes from the synthetic dataset. %
        The right column shows images from the real-world dataset.}
        \label{fig:dataset_examples}
    \end{center}
\end{strip}distinctiveness of the corresponding image region, possibly introducing outliers. %
In turn, this filtering adds areas with no estimates to the resulting depth maps, visible as black areas in depth maps in \Cref{fig:qualitative}. %
These holes, however, are typically filled again by fusing multiple depth maps into a consistent point cloud as described in the next section. %

\subsection{Fusion of depth maps into a joint point cloud}
For the fusion of our depth maps, we adopt the work of \citet{Whelan2015ElasticFusionDS,whelan2016elasticfusion} also known as ElasticFusion. %
Its main capabilities are tight tracking using a global model, registering RGBD images and performing global optimizations such as bundle adjustments and loop closure. %
Originally, this method was developed mainly for indoor use cases with close-range RGBD cameras like the Microsoft Kinect or the Intel RealSense. %
For the registration of RGBD images, ElasticFusion uses both a geometric procedure, which is based on the point-to-plane error, and a photometric technique using the intensities of the color images. %
In this context, the current RGBD image is compare with the back-projected view of the model seen from the pose of the previously fused frame. %
By jointly optimizing the combined cost volume, the current frame is transformed and can be integrated into the model. %


ElasticFusion relies heavily on acceleration from a graphics card for real-time processing. %
For this purpose, the representation and transformations of the global model are performed using OpenGL shaders. %
Other computationally intensive tasks like the \gls*{ICP} algorithm \citep{besl1992method} and also the cost volume optimization use implementations based on CUDA. %

In contrast to ElasticFusion, our focus lies on 3D reconstruction of objects from the aerial perspective of \glspl*{UAV}. %
The expected scenes differ greatly in the nature of the recording, the average depth, and, in terms of our datasets, the frame rate of the input videos. %
Since the scaling of our depth maps is conditioned by the scaling of the translation from ORB-SLAM2, the range of values varies considerably. %
However, since the scaling of ORB-SLAM2 remains in a similar range of values over different sequences, it is sufficient to scale the depth maps linearly. %

%% file: chapters/04_experiments.tex
\section{EVALUATION}
\label{sec:eval}
\sloppy

To evaluate our approach quantitatively, we not only compare the resulting 3D reconstruction with the ground truth, but also analyze the individual steps of our pipeline. %
This allows us to conclude on a potential error propagation, \eg caused by incorrectly estimated relative camera poses, which may influence the downstream depth estimation. %
For this purpose, we use a synthetic dataset of $51$ RGBD videos, which also includes extrinsic and intrinsic camera calibration. %
This enables us to evaluate the intermediate products of our pipeline, camera motion and depth maps. %
In addition to our synthetic dataset, we use real-world data with a ground truth created with the framework COLMAP  \citep{schoenberger2016sfm,schoenberger2016mvs}. %
\Cref{fig:dataset_examples} displays individual examples from our datasets. %

\Cref{fig:qualitative} displays qualitative examples from the two datasets used. %
The first two rows show exemplary input images of the sequences and the corresponding estimated depth maps. %
The next two rows represent the generated 3D model at the end of processing, once as colored point cloud and in the next row colored in yellow in comparison with the ground truth in blue. %
The last row shows the estimated trajectory compared to the ground truth flight path projected onto the $XY$ plane. %
The reference trajectory is represented as a dashed line and the metric deviation from it is highlighted in color. %

\begin{table*}
    \begin{minipage}{\linewidth}
        \begin{center}
            \setlength\tabcolsep{1pt} 
            \begin{tabular}{cccc}

                \rotatebox{90}{\small~~~~~~~Example image}&
                \includegraphics[width=52mm]{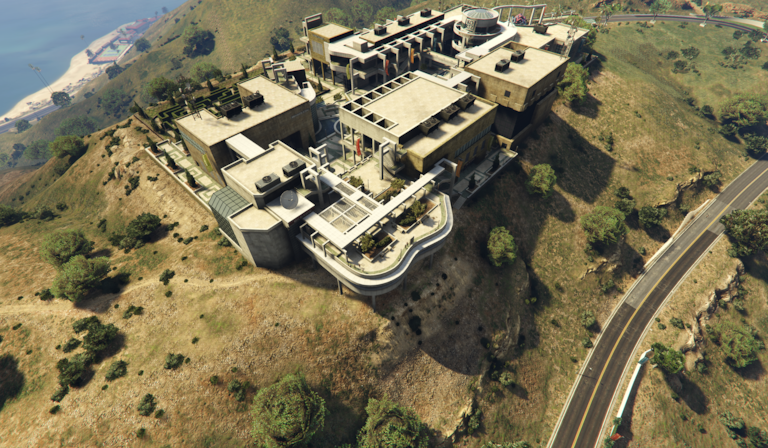}&
                \includegraphics[width=52mm]{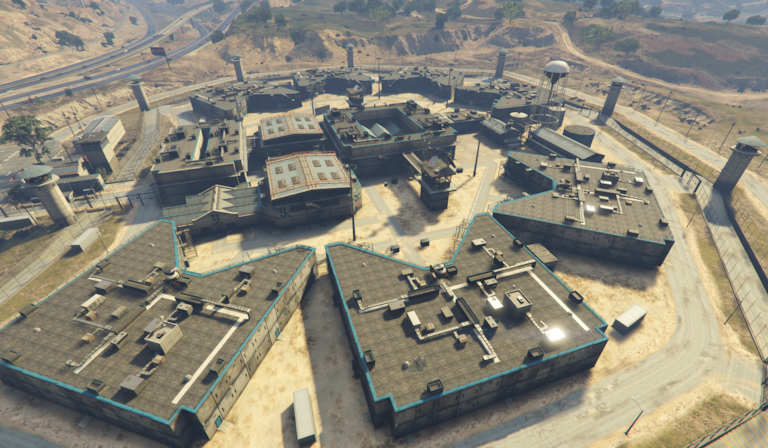}&
                \includegraphics[width=52mm,height=30.33333mm]{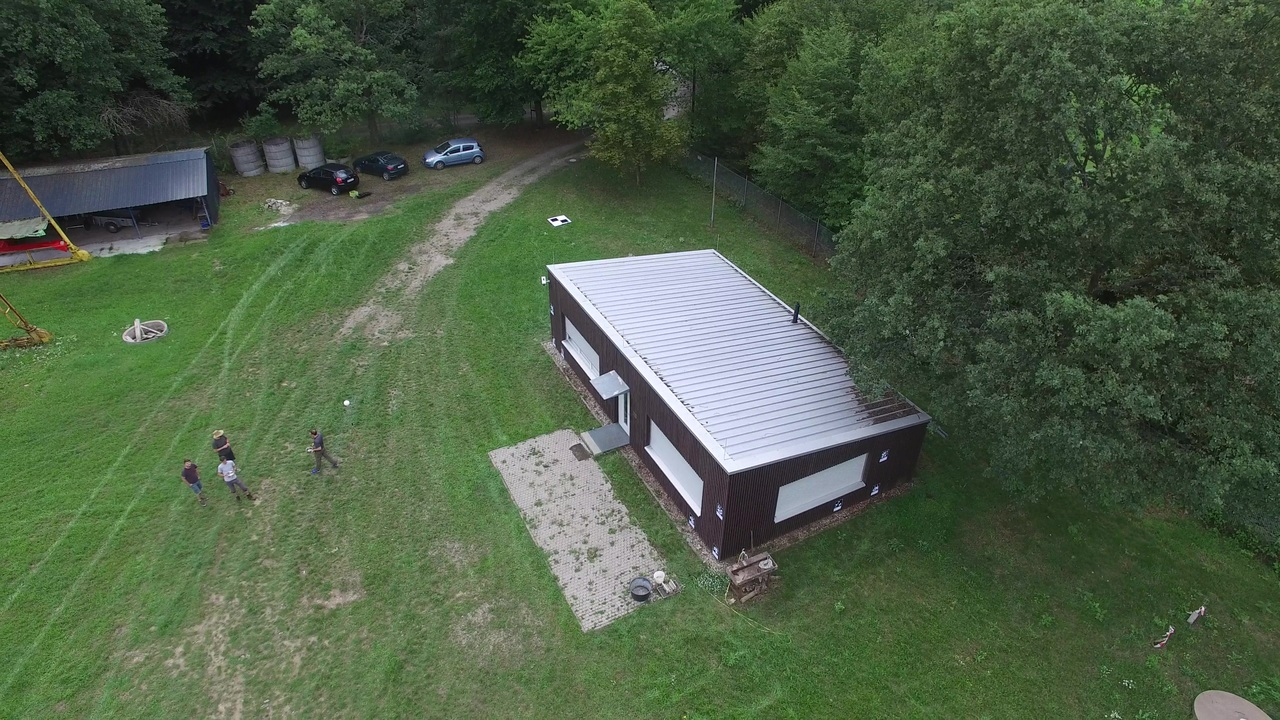}\\
            
                \rotatebox{90}{\small~~~~~~~~~~Depth map}&
                \includegraphics[width=52mm]{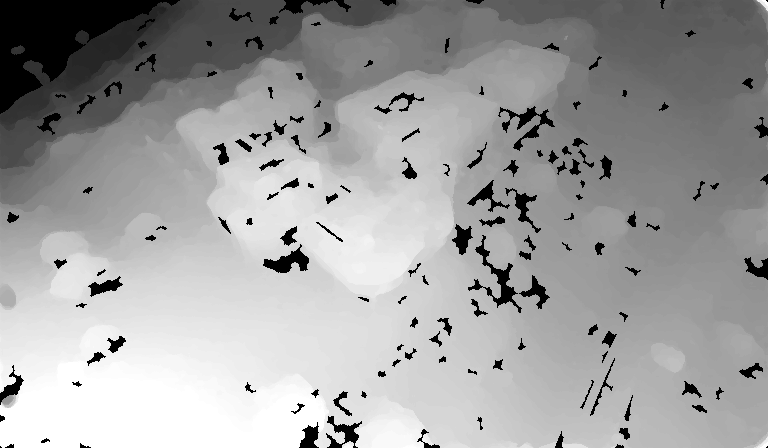}&
                \includegraphics[width=52mm]{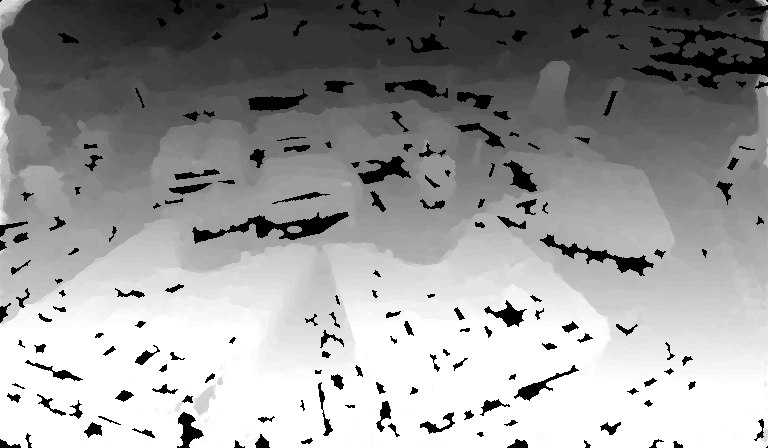}&
                \includegraphics[width=52mm,height=30.33333mm]{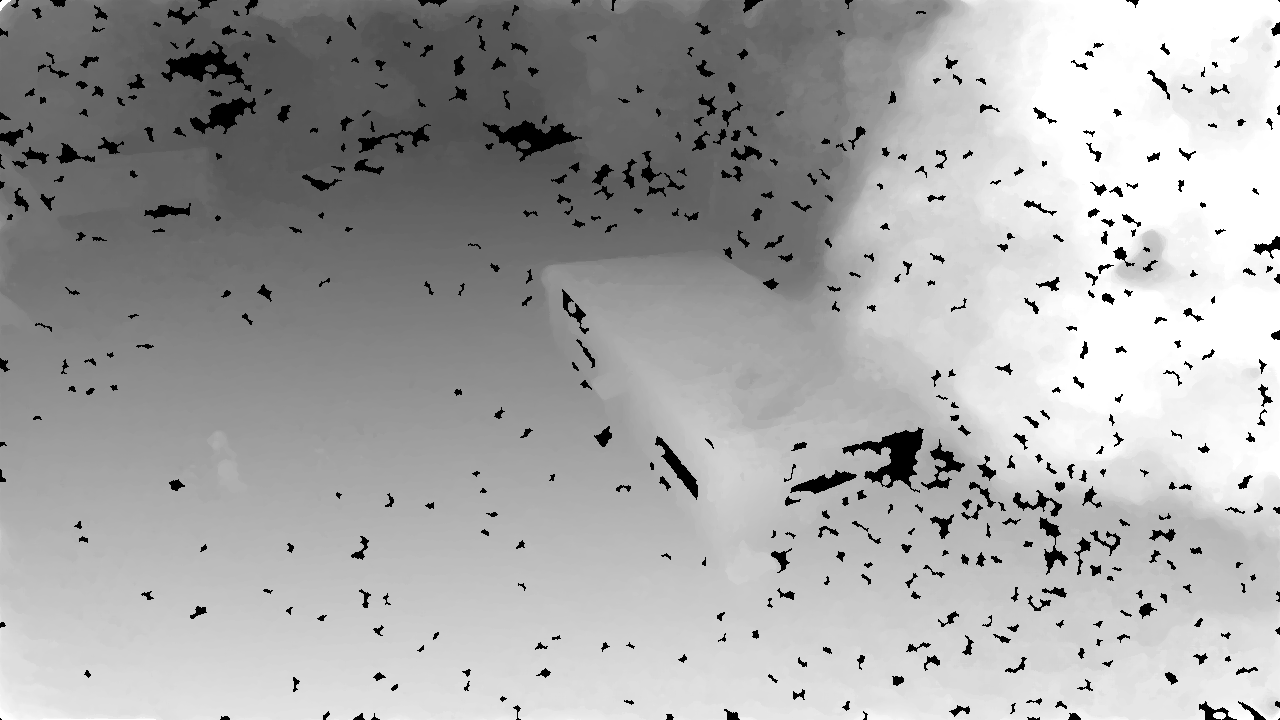}\\
            
                \rotatebox{90}{\small ~~~~~~~~~Point cloud}&
                \includegraphics[width=52mm,height=30.33333mm]{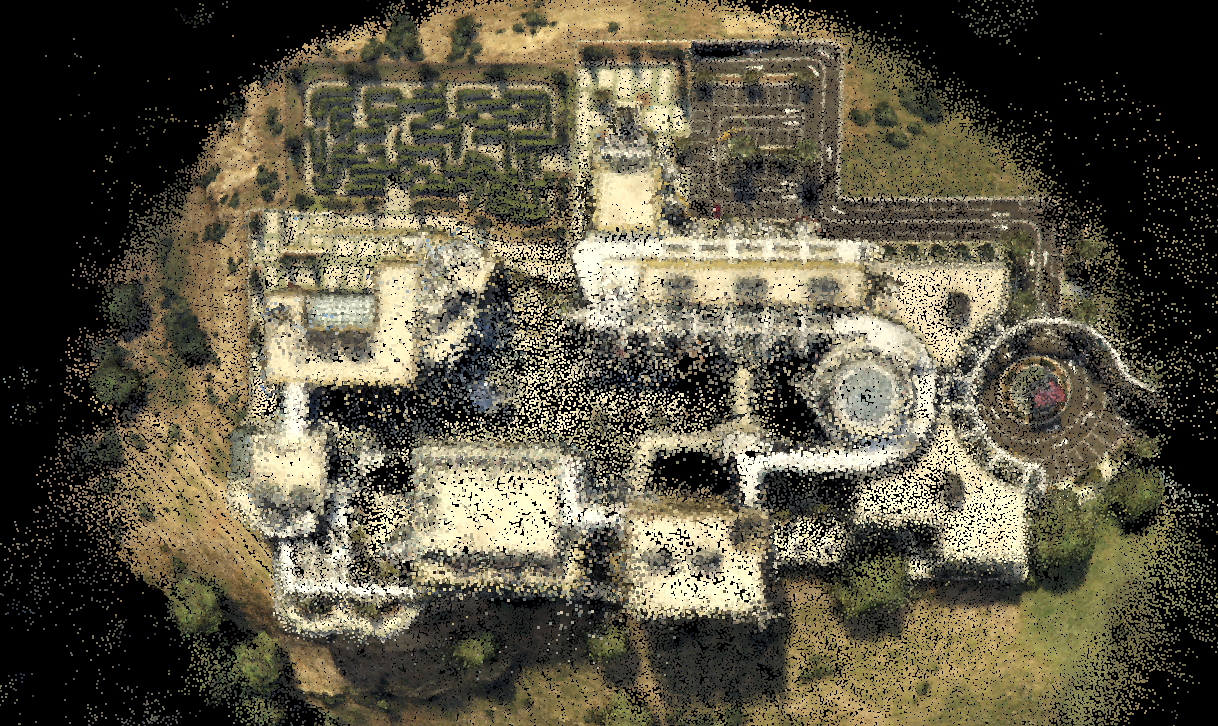}&
                \includegraphics[width=52mm,height=30.33333mm]{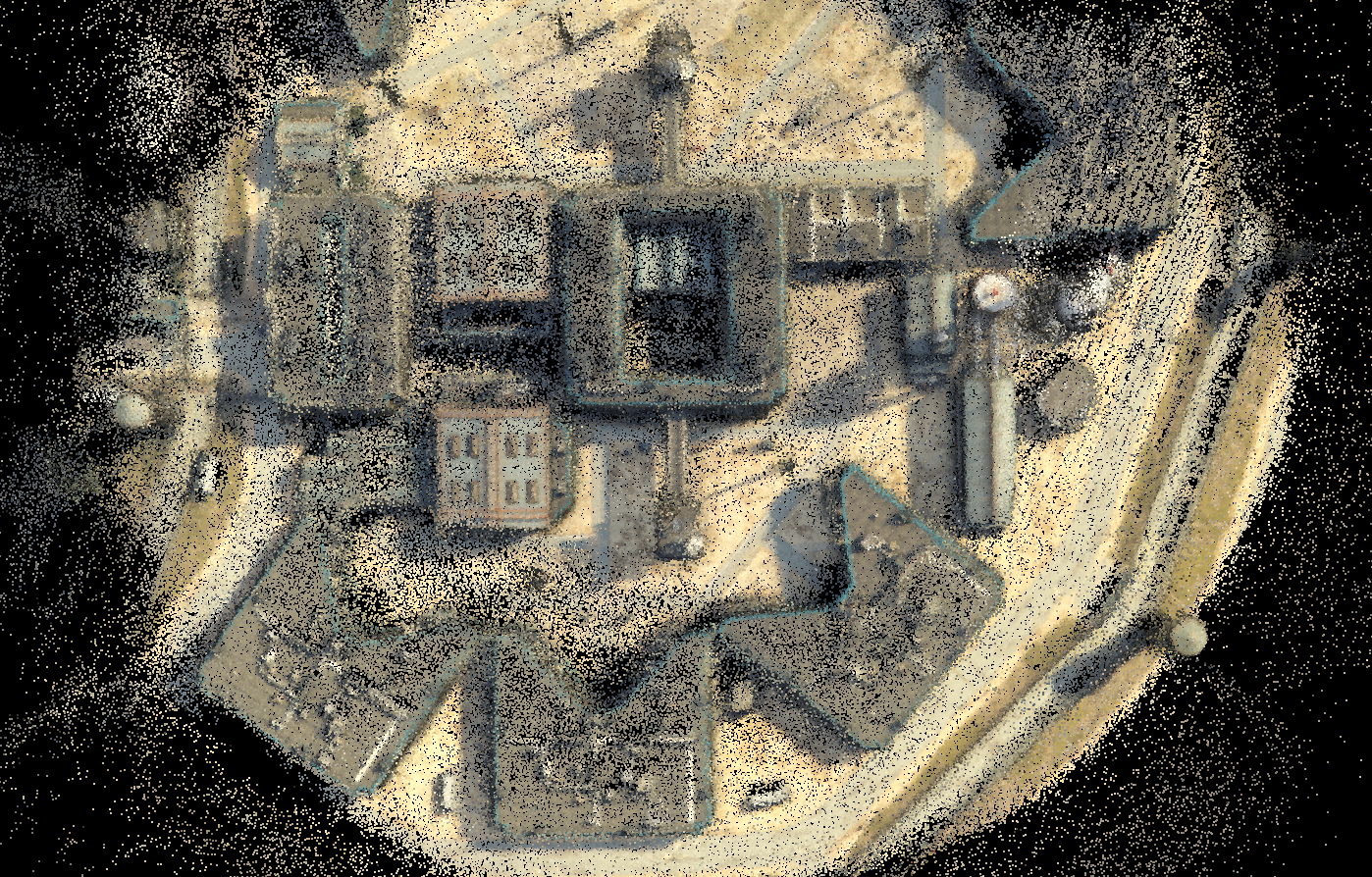}&
                \includegraphics[width=52mm,height=30.33333mm]{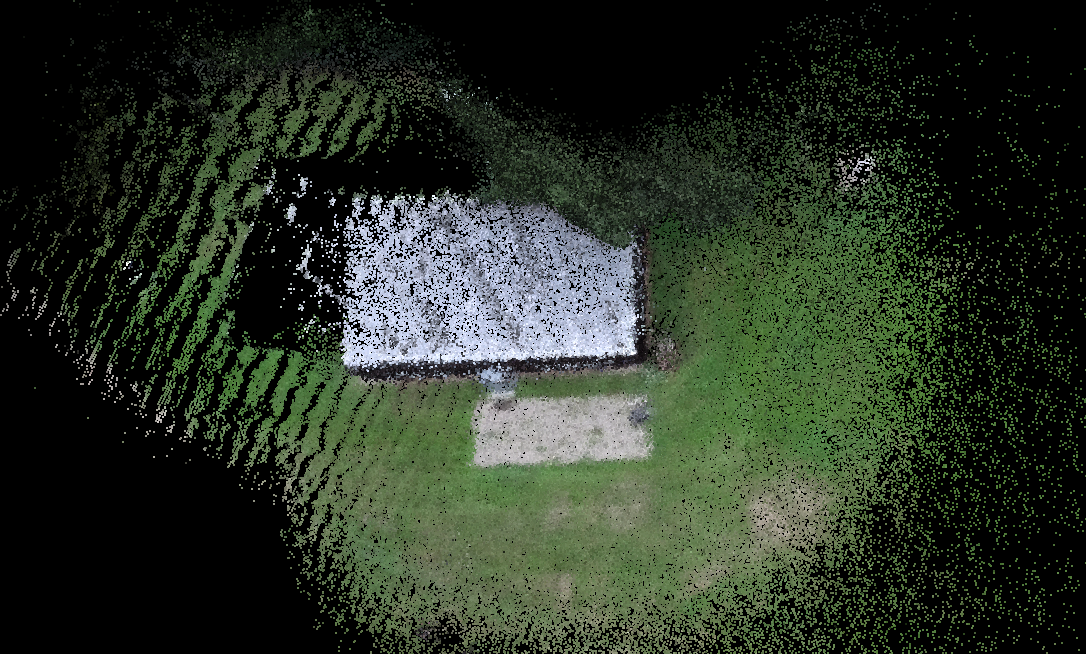}\\     

                \rotatebox{90}{\small Color coded point cloud}&
                \includegraphics[page=2,width=52mm,height=30.33333mm]{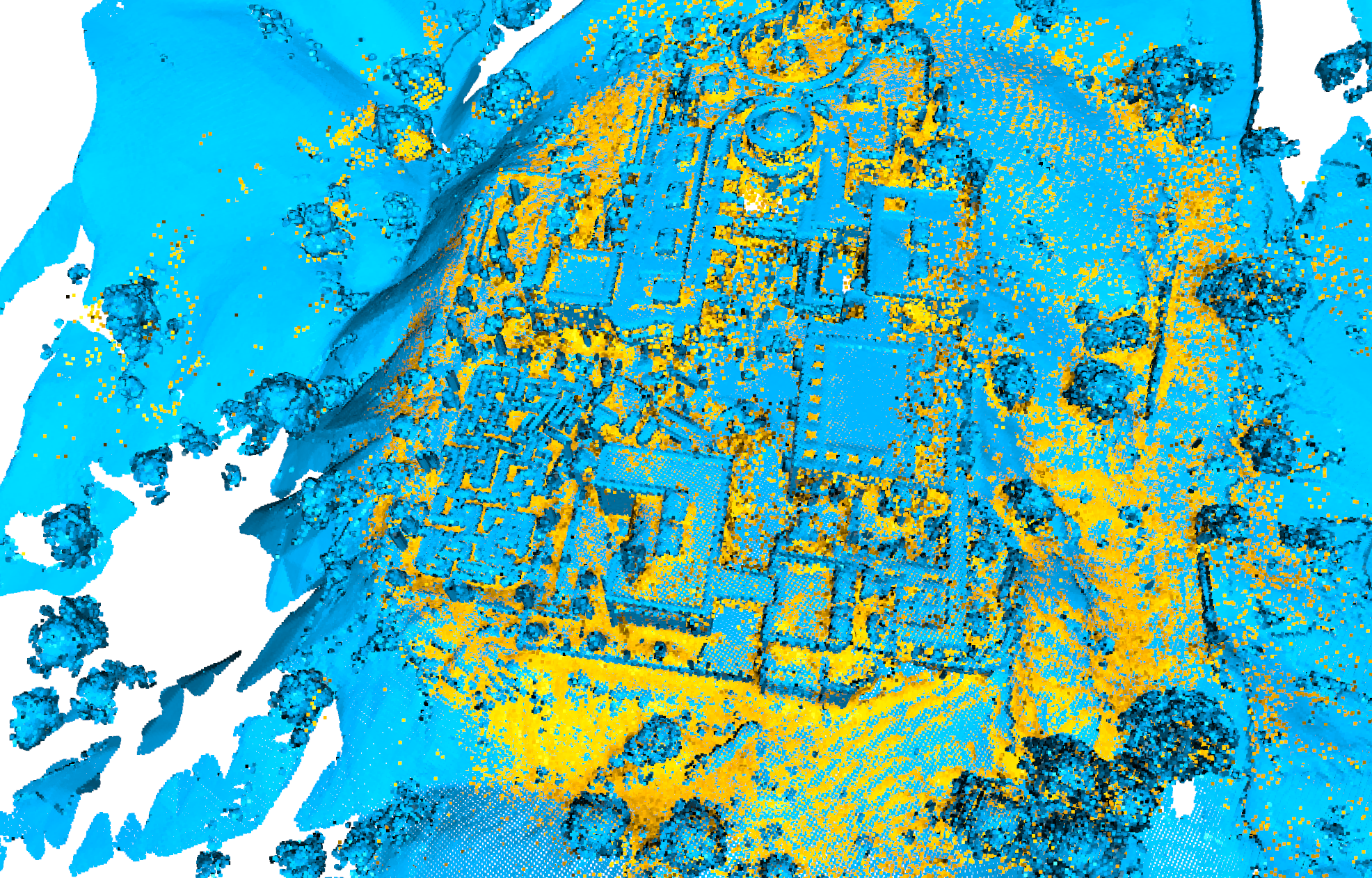}&
                \includegraphics[page=2,width=52mm,height=30.33333mm]{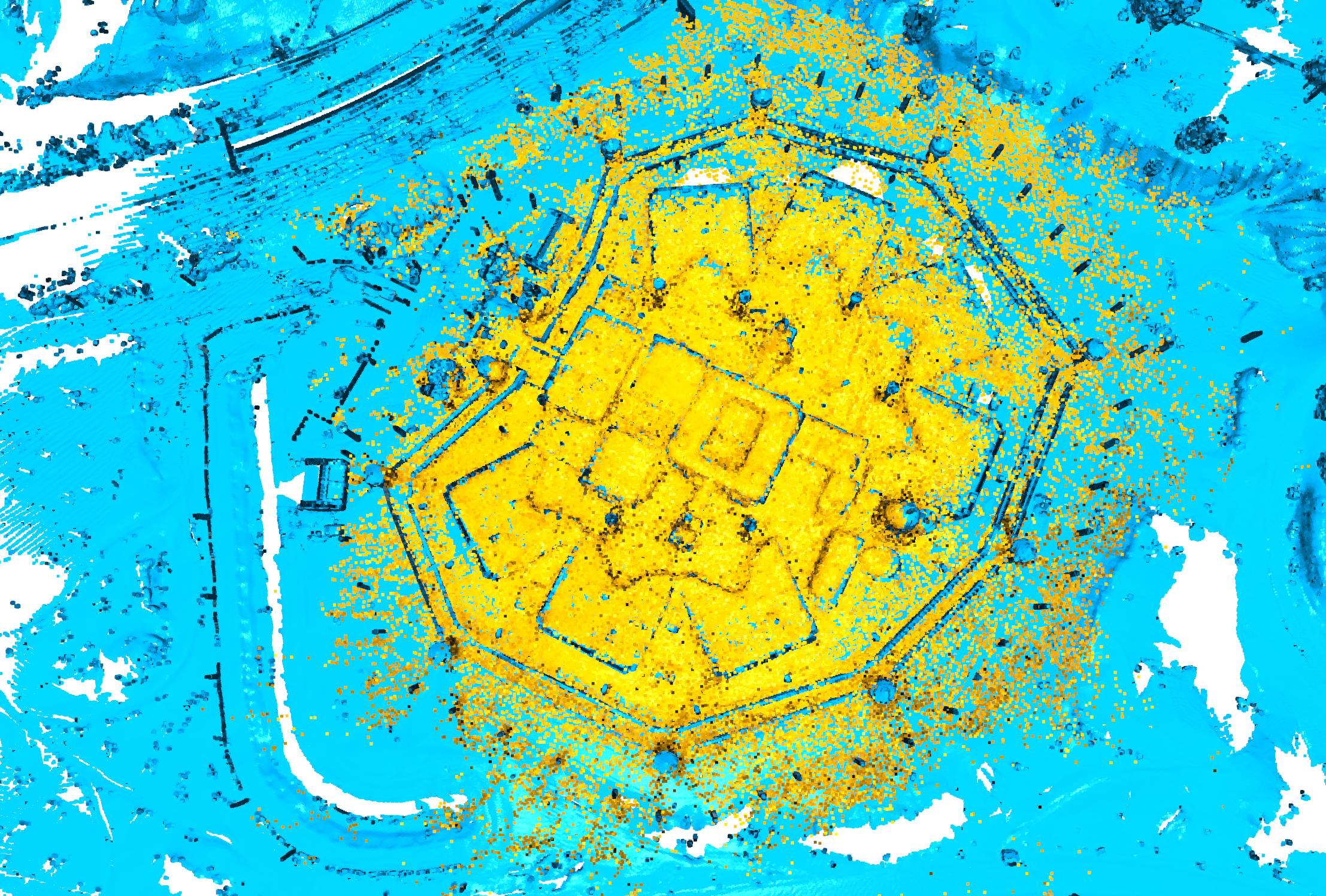}&
                \includegraphics[page=2,width=52mm,height=30.33333mm]{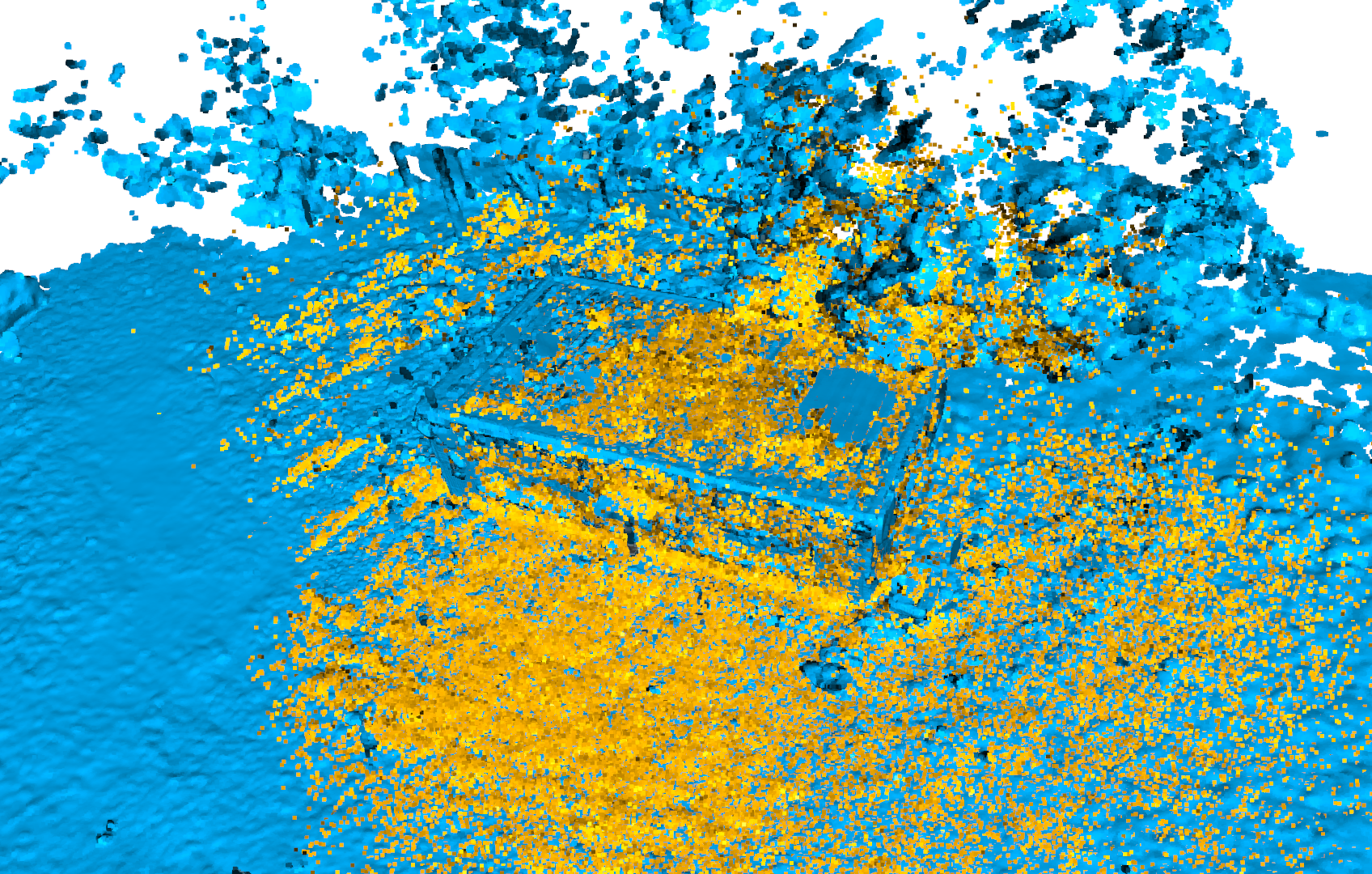} \\ 

                \rotatebox{90}{\small ~~~~~~~~~~~~~~~~~~~~~~~Camera trajectory}&
                \includegraphics[page=2,width=52mm,trim={60 45 21 24},clip]{figures/mansion/evo_ape_ef_xyz.pdf}&
                \includegraphics[page=2,width=52mm,trim={70 45 20 24},clip]{figures/jail/evo_ape_ef_xyz.pdf}&
                \includegraphics[page=2,width=52mm,trim={60 45 22 24},clip]{figures/building_15m/evo_ape_ef_xyz.pdf} \\ 
            
            \end{tabular}
            \small
            \captionof{figure}{Qualitative results on both datasets. %
            Rows 1 and 2 show exemplary input images with the corresponding estimated depth maps. %
            The following rows display the resulting colored point cloud after flight and the comparison with the ground truth point cloud. %
            The estimated point cloud is colored yellow and the ground truth blue. %
            The trajectory shown in the last row is the projection onto the $XY$ plane. %
            In this, the estimated trajectory is fitted to the ground truth trajectory using least squares due to the missing scale. %
            The resulting error is color coded.  }
            \label{fig:qualitative}
        \end{center}
    \end{minipage}
\end{table*}

\subsection{Datasets used for evaluation}
To evaluate our entire pipeline, we use two datasets for which the ground truth is obtained in different ways. %
For an extensive evaluation with respect to different scenarios in terms of flight altitude and scenery, we recorded a synthetic dataset using the video game \gls*{GTA}. %
In addition, we use a real-world dataset to validate the results. %
Both datasets consist of videos in oblique view, showing buildings from a low aerial perspective. %
In each case, the \gls*{UAV} flies around a central object and ensures that it is always in the center of the image. %
In the following, we describe the main properties of each dataset. %

\subsubsection{Synthetic dataset}
Our synthetic dataset consists of $51$ different sequences with a total of about $24{,}000$ images. %
These show scenes simulating the flight of a \gls*{UAV} orbiting different buildings.
For this, we use a modification of the code of \citet{johnson2016driving} to extract the data from the game engine. %
This allows to retrieve the depth maps and corresponding metadata of the current scene from the graphics card's memory. %

The sequences differ in altitude, flight speed, weather, time of day and scenery. %
For each extracted image, we store both a depth map and the position in the world coordinate system. %
This allows us to metrify our data, which simplifies the simulation and evaluation of realistic UAV flights, and makes the quantitative results easier to interpret. %
To generate ground truth 3D models from this, we back-projected all depth maps of a sequence into a joint point cloud. %
Due to dynamic objects like cars, pedestrians and vegetation, artifacts can occur in this way. %
For this reason, we remove points with a high average distance to their nearest $100$ neighbors with a threshold of $1.0$ standard deviation. %
This results in a very dense point cloud, which we reduce to a resolution of $0.01$\m for easier further processing.

\newcolumntype{C}{>{\centering\arraybackslash}X}%

\begin{table}[h] 
    \begin{tabular*}{\columnwidth}{@{\extracolsep{\fill}}|l|c|c|c|c|}
        \hline
        \rule{0pt}{\normalbaselineskip}
        &Flight height&Angular&Fps&Velocity\\   
        &in \m&velocity in r/s&&in \m/s\\  
        \hline
        \rule{0pt}{\normalbaselineskip}%
        $\mu$&71.10&0.0156&2.33&1.38\\
        $\sigma$&30.88&0.0114&0.47&0.67\\
        min&23.00&0.0000&0.22&0.00\\
        max&134.00&0.0935&3.98&4.77\\
        \hline
    \end{tabular*}
    \small   
    \caption{Key information for our synthetic dataset. }
    \label{tab:gta_facts}
\end{table}

\Cref{tab:gta_facts} shows the most important characteristics for our synthetic dataset. %
Since we do not have direct access to the flight altitude, we can only retrieve the position in the reference coordinate system. %
For this reason, we approximate it by subtracting the $5$\% quantile of the vertical coordinates of the ground truth point clouds from the position of the camera for each sequence. %
In this way, we record an average flight altitude of $71.1$\m which appears consistent with the mean depth of $83.4$\m for each depth map. %
The maximum possible frame rate depends on the speed at which the data can be extracted from the graphics card's memory. %
In our case, the frame rate of our videos fluctuates around a mean value of $2.33$ \fps at an average flight speed of $1.381$\m/s and a mean angular velocity of $0.0156$ radians per second. %

\subsubsection{Real-world dataset}
To validate our results on real-world data, we also use a dataset recorded by a  DJI Phantom $3$ Professional. %
Here, we are dealing with a rural dataset, where a small \gls*{UAV} is used to fly around buildings. %
The dataset consists of $4$ sequences around $2$ buildings, with in total $1{,}500$ images. %
In contrast to the synthetic dataset, the flight altitude ranges from about $8$\m to $15$\m and the frame rate of the recording is $10$ \fps. %
Due to the fact that the \gls*{UAV} is controlled by hand and does not follow a strictly orbital flight path, the trajectories are much more unsteady. %
The ground truth is generated with the photogrammetry framework COLMAP. %
In this way, data for intrinsic and extrinsic camera calibration are computed, as well as depth maps and 3D models in the form of dense point clouds. %
To metrify the models created with COLMAP, we use GPS metadata from a double grid flight whose images are coregistered into the 3D models. %
In this way, the point cloud is scaled into a local ENU frame, and subsequently flight trajectories, depth maps, and dense point clouds are transformed. %
Due to the low altitude and camera angle, the mean depth is $11.5$\m. %

\subsection{Camera tracking and depth map registration}\label{sec:pose_eval}
In the presented approach, we estimate camera motion at two points of our processing pipeline: %
first locally between adjacent images using ORB-SLAM2 and then between the estimated depth map and the global model during registration. %
Due to the fact that our poses from ORB-SLAM2 are only locally consistent, but mostly not yet globally optimized at this point, a trajectory that simply aggregrates these has a high drift. %
This behavior is visible in \Cref{fig:ef_orb_traj} and stands in contrast to the final estimated trajectory after registration with the model. %
It should be noted, that after our alignment with the ground truth, the $Z$-axis points upwards and not forward as it normally does when ORB-SLAM2 is initialized. %

Since our ground truth is given in meters, but our estimated trajectories do not have an interpretable scale, it is necessary to align them. %
For this, we use the software library evo \citep{grupp2017evo}. %
Using an approach based on least squares, the estimated trajectory is thereby translated and scaled uniformly. %
Since our local poses computed with ORB-SLAM2 are by nature inaccurate at the global scale and globally optimized poses are not available during processing, we only evaluate the trajectory after registration of the depth maps. %
From the $51$ sequences, $47$ provide useful values with our parameterization. %
For the rest, the registration fails, leading to a success rate of $92.15$\%. %
We exclude the sequences which have failed from the following evaluation. %

\begin{minipage}{\linewidth}
    \begin{center}
        \includegraphics[width=70mm]{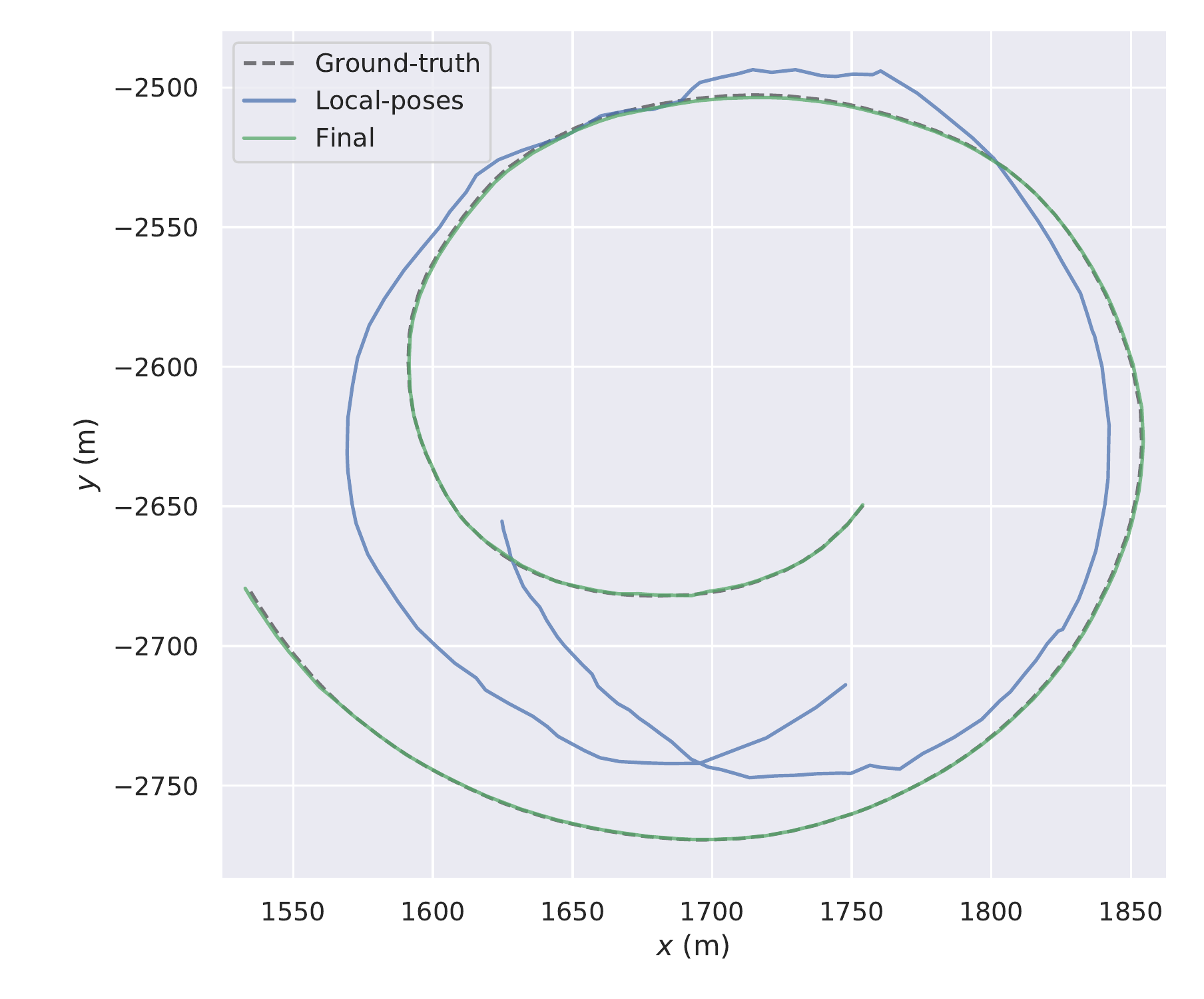}
        \small 
        \captionof{figure}{Example for the drift of ORB-SLAM2 if the local poses are only aggregated. %
        The graph shows the projection onto the $XY$ plane, with the dashed line representing the ground truth trajectory. %
        In addition, the aggregated local poses from ORB-SLAM2 are plotted in blue and the final trajectory after registration against the model is plotted in green. %
        }
        \label{fig:ef_orb_traj}
    \end{center}
\end{minipage}

\begin{table}[h] 
    \begin{tabularx}{\columnwidth}{|p{1.5cm}|C|C|C|}
        \hline
        \rule{0pt}{\normalbaselineskip}%
Dataset&RMSE&\acrshort*{MAE}&$\sigma$\\
        \hline
        \rule{0pt}{\normalbaselineskip}%
Synthetic&1.481&1.235&0.800\\
Real-world&0.577&0.519&0.251\\
        \hline
    \end{tabularx}
    \small
    \caption{Results for trajectory estimation after registration with the model.}
    \label{tab:results_pose}
\end{table}

As long as the tracking of the camera movement does not fail completely, our generated 3D models hardly show any drift. %
This is visible in both the point clouds and the camera trajectories and is confirmed by the quantitative results, since we could not find any correlation in our investigations between duration of flight and magnitude of the error. %
Even relatively long sequences with over $1{,}800$ images are tracked consistently, although we do not perform any post-processing steps to optimize our trajectories. %
Accordingly, the drift of the ORB-SLAM2 poses is fully compensated by the fusion into a single joint 3D model. %
It is probably helpful that our scenarios mainly consist of flying around individual buildings in oblique view and not of simple overflights. %
\Cref{tab:results_pose} shows the results for camera tracking on both datasets. %
We attribute the higher error on the synthetic dataset to the more complex scenes with higher flight altitude and duration. %

\subsection{Depth map estimation}
In addition to evaluating camera tracking, we also investigate the second stage of our processing chain, which is the estimation of depth maps. %
Since this requires the relative camera motion between the five sampled images $I_i$ to estimate depth, we also indirectly evaluate ORB-SLAM2's estimation of local poses when evaluating our depth estimation. %
By comparing the results with experiments using ground truth poses, we can correlate the performance with the camera tracking by ORB-SLAM2. %

The scaling of our depth maps depends on the estimation of the camera's translation and therefore does not have a directly interpretable range of values. %
To compare our estimates with the metric ground truth, we use median scaling to align the different value ranges. %
In case of our synthetic dataset, we limit the maximum depth to $300$\m. %

\begin{table}[h] 
    \begin{tabularx}{\columnwidth}{|p{1.5cm}|C|C|C|C|}
        \hline
        \rule{0pt}{\normalbaselineskip}%
        Dataset			&	\acrshort*{RMSE} & \acrshort*{MAE} &	$\delta_{1.25}$	&	$\delta_{1.05}$	\\        
        \hline
        \rule{0pt}{\normalbaselineskip}%
        Synthetic&11.406&0.034&0.974&0.876\\
        Real-world&0.966&0.035&0.984&0.831\\
        \hline
    \end{tabularx}
    \small   
    \caption{Results of our depth estimation on both datasets.}
    \label{tab:results_depth}
\end{table}

The results listed in \Cref{tab:results_depth} for the real-world dataset are comparable to results based on ground truth camera poses. %
However, our data reveals a $6.6$\% drop in $\delta_{1.05}$ accuracy, which we attribute to the less accurate camera poses by ORB-SLAM2 and poorer image selection. %
For the most part, the results for the synthetic dataset are consistent with those for the real one. %
However, the \acrshort*{RMSE} is significantly higher. Presumably, this is due to the much higher depth range of these scenes, which might lead to higher squared errors and in turn a higher \acrshort*{RMSE}. %
In addition, dynamic objects such as cars or pedestrians lead to artifacts in the depth maps. %

\subsection{Evaluation of our 3D reconstruction}
As mentioned before, the generated 3D models do not have an interpretable scale. %
In order to compare our results with the ground truth we have to align the value ranges first. %
Due to the fact that the two point clouds that are to be matched have very different properties with respect to size, density, rotation and translation, a direct alignment with ICP did not prove to be successful. %
Because the maximum range of our depth maps is smaller, our 3D model also covers a smaller area than the ground truth. %
Instead, we use the following three steps: %
1) Rough alignment using the similarity transformations of the camera trajectories from \Cref{sec:pose_eval}. %
2) Finer adjustment with classic ICP using the point-to-plane error metric. %
3) Using ICP with scaling for the final result. %
As error measure we use the \gls*{RMSE} of the individual points to their nearest counterpart. %
For both error calculations as well as for alignment and visualization of our point clouds, we use the framework Open3D \citep{Zhou2018}. %

\begin{table}[h] 
    \begin{tabularx}{\columnwidth}{|p{1.5cm}|C|C|}
        \hline
        \rule{0pt}{\normalbaselineskip}%
Dataset&\acrshort*{RMSE}&$\sigma$ \acrshort*{RMSE} \\           
       \hline
       \rule{0pt}{\normalbaselineskip}%
       Synthetic&0.744&0.616 \\
       Real-world&0.141&0.033 \\
        \hline
    \end{tabularx}
    \small
    \caption{Results for the comparison of our 3D models with the ground truth.}
    \label{tab:results_icp}
\end{table}

Similar to the case of pose and depth estimation, the \acrshort*{RMSE} of the 3D reconstruction for our synthetic dataset is considerably higher than that for the real-world dataset. %
However, the standard deviation of the \acrshort*{RMSE} of $0.616$ indicates that there are large differences between individual sequences. %
This is also visible in the varying point cloud spacing of the 3D models. %
Depending on the flight altitude of the sequence, these range from $0.02$\m to $0.46$\m. %
As displayed in \Cref{fig:qualitative}, our 3D models reproduce the geometry of the scene well, but contain significantly more noise compared to the ground truth. %
This is especially visible in the fourth row of the figure, as our models appear much thinner because their surface is partially below that of the ground truth. %

\subsection{Speed of execution}
We perform the experiments with respect to execution speed on a desktop computer with a NVIDIA GTX 1080ti GPU. %
Both depth estimation and fusion are accelerated by CUDA. %
Depending on the resolution of the images, both ORB-SLAM2 and ElasticFusion already run in real-time. %
In contrast, our depth estimation requires a mean of $220$\ms at a resolution of $768\times448$ pixels, which is not real-time capable. %
But since we do not execute this for every frame, we can still keep up with $30$\fps input video streams. %
However, our approach depends on how many keyframes are sampled by ORB-SLAM2 and which of them are suitable for depth estimation. %
For example, at higher flight speeds, ORB-SLAM2 will select new keyframes at a higher frequency, as they provide novel image content more often. %
In order to keep up with the video stream in scenes with a large number of image constellations suitable for depth estimation, we discard images when our depth estimation is still in progress. %

In our experiments, we obtain an average subsampling ratio of $2.03$\% on our synthetic dataset. %
This means, that after the keyframe selection by ORB-SLAM2 and our search for images suitable for depth estimation, in the end only $2.03$\% of the original input images are used for depth estimation and fusion. %
However, some frames are indirectly included in the depth estimation as supplementary views and the rate is higher during the actual active phase, since a sufficient number of keyframes must first be collected at the beginning of each sequence. %



\subsection{Failed registration and broken model}\label{sec:broken}
As described in \Cref{sec:pose_eval}, our approach does not yield a useful result on $7.85$\% of the synthetic sequences. %
This is mainly due to the fact that the registration of depth maps fails because of challenging scenery. %
If this happens once with subsequent correct registration, this will not be critical since these outliers can be eliminated over time by a sufficient number of correct values. %
Repeated incorrect registrations result in a very noisy model, which makes subsequent correct registrations even more difficult. %
Since our approach uses a global model to integrate all depth maps, there is currently no way to recover from this state. %

In other cases, such as rapid camera movements, if depth maps cannot be estimated over a certain period of time, it is difficult to register the next depth map into the global model due to insufficient overlap. %
We therefore depend on the sampling of images suitable for depth estimation at regular intervals. %
Our experiments have also shown that scenes with particularly flat terrain and few structures present great difficulties for the registration. %

Depending on the application scenario and the type of depth estimation, softening the sampling criteria may also be an option. %
For this purpose, the capabilities of the depth estimation must be known beyond the optimal parameters in order to determine up to which limit acceptable values are still generated. %
In this case, one would prioritize further model generation and accept inferior depth maps with accompanying noise in the 3D model. %
If quality is more relevant, it should be detected in real-time that the model has an error, \eg to adjust further flight parameters. %

\subsection{Error propagation across the processing steps }
In order to analyze the dynamics of our processing chain and to identify a potential error propagation, we investigate if there are direct dependencies between the individual steps. %
Depending on the sequence, the \gls*{RMSE} of our point clouds is a poor criterion for a good 3D model. %
Especially in scenes with a lot of flat terrain, models where the registration of the depth maps fails also deliver good results as long as the ground plane is well aligned. %
On the other hand, for the detection of completely broken models, the analysis of the camera trajectories is much more meaningful. %

Conversely, a high \gls*{RMSE} of the camera trajectory does not correspond to a high \gls*{RMSE} of the point cloud. %
However, in contrast, the \gls*{RMSE} of the point clouds increases when the quality of the depth maps decreases. %
This corresponds to the qualitative impression of more noisy point clouds in sequences with particularly low accuracy in depth estimation. %
The fact that we do not observe a similar drop in performance in camera tracking suggests that it is relatively robust against outliers. %

%% file: chapters/06_conclusion.tex
\section{CONCLUSION \& FUTURE WORK}
\label{sec:conclusion}

\sloppy


In this paper, we present an end-to-end approach for real-time 3D model generation from aerial imagery captured from small versatile UAVs. The evaluation shows that our framework allows creating 3D models from diverse aerial images in real-time. Although the depth estimation itself is not real-time capable, our pipeline can easily keep up with 30 fps video streams. This is because not every frame of the input video provides new information and is chosen for our depth estimation. For this purpose, our method selects images from the video stream that satisfy geometric constraints.

Some of the challenges discussed in this work are also known as the \gls*{NBV} problem. %
This deals with the problem of determining optimal measurement locations in online processing and describes the conflict of whether to settle for the current non-optimal sensor position or to wait for a potential better one. %
In the future, we want to address this issue better by more closely integrating our individual steps in the processing chain and linking them with potential feedback loops. %
Metrification of the SLAM algorithm via \gls*{GPS} or \gls*{IMU} could also ease the dynamic selection of the optimal baseline for \gls*{MVS}, as well as the sampling rate for our depth estimation. %
